# Information-Theoretic Framework for Self-Adapting Model Predictive Controllers

Wael Hafez
Semarx Research LLC
Alexandria, VA, USA
w.hafez@semarx.com

Amir Nazeri
Semarx Research LLC
Alexandria, VA, USA
amir.nazeri@semarx.com

*Abstract*—Model Predictive Control (MPC) is a vital technique for autonomous systems, like Unmanned Aerial Vehicles (UAVs), enabling optimized motion planning. However, traditional MPC struggles to adapt to real-time changes such as dynamic obstacles and shifting system dynamics, lacking inherent mechanisms for self-monitoring and adaptive optimization. Here, we introduce Entanglement Learning (EL), an information-theoretic framework that enhances MPC adaptability through an Information Digital Twin (IDT). The IDT monitors and quantifies, in bits, the information flow between MPC inputs, control actions, and UAV behavior. By introducing new information-theoretic metrics we call entanglement metrics, it tracks variations in these dependencies. These metrics measure the mutual information between the optimizer's input, its control actions, and the resulting UAV dynamics, enabling a deeper understanding of their interrelationships. This allows the IDT to detect performance deviations and generate real-time adaptive signals to recalibrate MPC parameters, preserving stability. Unlike traditional MPC, which relies on error-based feedback, this dual-feedback approach leverages information flow for proactive adaptation to evolving conditions. Scalable and leveraging existing infrastructure, this framework improves MPC reliability and robustness across diverse scenarios, extending beyond UAV control to any MPC implementation requiring adaptive performance.

I. INTRODUCTION

Model Predictive Control (MPC) is a cornerstone in unmanned aerial vehicle (UAV) motion planning, enabling precise trajectory optimization while maintaining adherence to system constraints. MPC enhances safety, navigation accuracy, and obstacle avoidance in complex and dynamic environments by predicting future states and dynamically adjusting control signals [1]. Advances in real-time optimization techniques have further expanded the capabilities of MPC [2], establishing it as an essential tool for achieving robust autonomous flight [3].

However, despite its proven strengths, MPC encounters significant challenges in environments characterized by uncertainties and disturbances. These include unmodeled dynamics, external forces such as wind gusts [1], and interactions with other agents [4]. While robust MPC approaches address such issues by adapting to changing conditions and maintaining stability, they remain constrained by operational limitations [5]. Robust optimization [5], adaptive control [6], and learning-based methods [7], improve resilience but face trade-offs. Robust optimization methods often rely on conservative uncertainty estimates, limiting responsiveness to rapidly evolving conditions [8][9][10]. Similarly, adaptive control techniques require extensive tuning to handle complex, nonlinear behaviors, and learning-based methods demand substantial computational resources and large datasets, often exceeding UAV system capacities [7][8].

We introduce an information-theoretic framework that enhances Model Predictive Control (MPC) adaptability by quantifying mutual information within the control loop. This approach measures how effectively MPC aligns its control decisions with system dynamics using information- and entropy-based metrics. Specifically, we compute the mutual information $MI(S,A;S')$, where $S$ represents the system state (MPC inputs), $A$ denotes the control actions, and $S'$ is the resulting next state. This quantifies how well knowledge of states and actions predicts future system behavior, serving as a real-time measure of model-reality alignment.

*A. Dual-Feedback through the Information Digital Twin*

To quantify the alignment between control decisions and system dynamics, we measure mutual information $MI(S,A;S')$ using an entanglement-based approach integrated into a dual-feedback framework. This design enables continuous monitoring and real-time adaptation whenever the alignment between the control model and physical reality degrades.

The dual-feedback architecture (illustrated in Fig. 1) consists of two nested loops. The inner loop performs standard Model Predictive Control (MPC) tasks: evaluating the system state, optimizing trajectories, and generating control actions. The outer loop, implemented through an Information Digital Twin (IDT), monitors the mutual information between MPC inputs (e.g., reference trajectories), control actions, and system responses.

In contrast to traditional MPC, which lacks built-in mechanisms for self-assessment [11], the IDT leverages entanglement metrics to detect discrepancies between the control model and actual system behavior. When these deviations exceed a predefined threshold, the IDT generates adaptive signals to recalibrate MPC parameters, ensuring stability without necessitating a complete model overhaul. This approach significantly enhances resilience under dynamic conditions [12], making it especially suited for applications like UAV navigation in unpredictable environments [13].

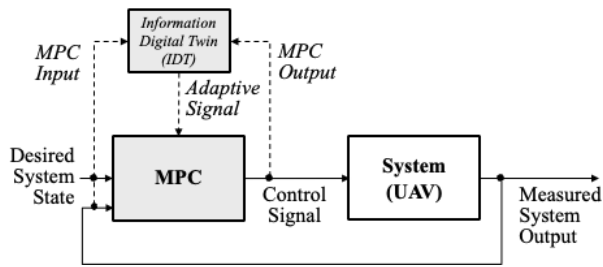

Fig. 1. Schematic of the dual-feedback architecture for adaptive MPC. The inner loop (solid lines) computes control actions by comparing reference inputs with system outputs. The outer loop (dotted lines), driven by the IDT, assesses input-output relationships to produce adaptive signals, enabling real-time adjustments while maintaining stability.

### B. Theoretical Foundation and Computational Feasibility

This paper defines the mathematical framework of our approach and evaluates its computational feasibility. Although experimental validation is absent, we provide a rigorous analysis of implementation requirements and links to control theory. Entanglement calculations, based on probability distributions over discretized state-action spaces, require $O(N)$ operations for N bins and run parallel to the main MPC optimization, keeping computational overhead low and timing impact minimal.

Our entanglement metrics offer an information-theoretic complement to traditional control measures. Unlike observability (state inference from outputs) and controllability (state reachability), entanglement assesses how well the controller's model reflects real input-output dynamics. By focusing on information flow rather than error tracking, these metrics detect model-reality misalignment early, enabling proactive adaptation before performance noticeably declines.

## II. RELATED WORK

Early efforts to weave information theory into control systems, such as Saridis' entropy-based methods [14], and Feldbaum's dual control [15], established key foundations but were limited by computational constraints. These ideas shaped stochastic control, though their complexity restricted widespread use.

Modern adaptive Model Predictive Control (MPC) methods have since emerged to address uncertainty. Robust MPC [16], focuses on worst-case scenarios, often at the expense of responsiveness, while adaptive MPC [17], requires precise, application-specific tuning. Learning-based MPC [18], employs data-driven techniques but faces challenges with interpretability and data dependency, rendering these methods domain-specific and lacking universal metrics.

Information-theoretic approaches have further advanced the field. Entropy has been applied to quantify unpredictability [19], and mutual information to assess control performance [20]. Touchette and Lloyd [21] [22], pioneered this perspective, optimizing control through information flow, though their focus remains on static or pairwise relationships. Chernyshov [1] identified a critical gap: these methods rarely tackle dynamic, multi-component interactions or real-time adaptability.

Practical implementations like traditional dual-feedback architectures, such as dual-loop control [24], manage disturbances but do not directly monitor information patterns. Similarly, digital twins [25] replicate physical behavior for monitoring and prediction in control applications, offering significant value. However, they typically overlook information flow dynamics. Our Information Digital Twin (IDT), by contrast, enhances this concept by tracking real-time entanglement metrics, enabling early detection of system misalignments before performance degrades.

While prior information-theoretic approaches to control have explored entropy and mutual information, they typically lack concrete methods for computing these quantities in continuous control systems. Our method addresses this gap by discretizing state, action, and outcome spaces (i.e., dividing continuous values into meaningful intervals, or "bins," for statistical analysis), enabling real-time computation of probability distributions and entropies from operational, real-time data. This discretization not only allows precise measurement of information relationships but also produces differentiated signals that guide targeted control adjustments. By mapping continuous variables into physically meaningful bins, we enable practical adaptation based on evolving dependencies across states, actions, and outcomes—capabilities absent from earlier, largely theoretical frameworks.

## III. INFORMATION AND ENTANGLEMENT BACKGROUND

This section lays the information-theoretic groundwork for Entanglement Learning (EL) by viewing control systems, such as Model Predictive Control (MPC), from a communication perspective. In this view, MPC generates control actions that shape statistical dependencies between MPC inputs, control actions, and UAV responses. Control systems inherently reduce uncertainty in system behavior, making information theory a natural lens for analyzing their performance. Our goal is to quantify these relationships using information-theoretic metrics and demonstrate their potential to maintain MPC alignment under dynamic conditions.

We begin by outlining information as defined by communication theory, then explore how it captures MPC's role in sustaining dependencies across state transitions. Finally, we adapt entanglement—inspired by quantum mechanics—to measure multiple information relationships across all MPC dynamics.

### A. Information as Uncertainty Reduction

Shannon defined information as the reduction of uncertainty about one variable given knowledge of another [26]. Measured in bits, it quantifies how much knowing a message from a source reduces uncertainty about the signal received at a destination. In communication systems (Fig. 2, top diagram), this process involves source coding to represent message statistical characteristics, channel coding to counter uncertainty during transmission, and decoding at the destination to recover the original message despite communication errors. This framework fundamentally relies on statistical dependencies between variables and prior distributions reflecting initial uncertainty.

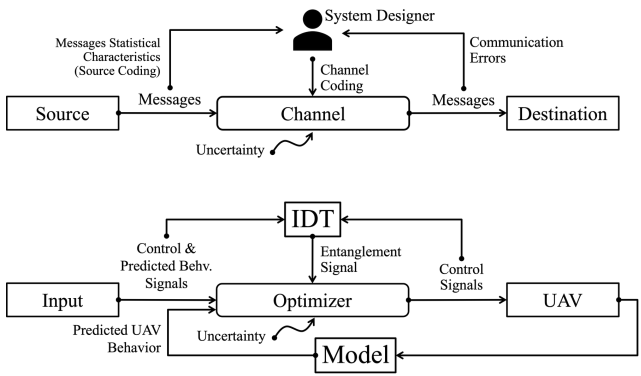

Fig. 2. Communication and Control System Parallels. The top diagram depicts a traditional communication system where a human designer establishes fixed source and channel coding based on message statistics and expected noise. The bottom diagram illustrates our EL approach, where the Information Digital Twin (IDT) replaces the human designer, continuously monitoring information patterns between control inputs and system behavior to provide adaptive signals to the optimizer. This parallel highlights the shift from static, designer-defined relationships to dynamic, continuously measured information dependencies.

### B. Control Systems as Communication Channels

Control systems like MPC parallel this communication model (Fig. 2, bottom diagram), [21]. The control inputs serve as the source, generating messages (predicted behavior signals) that pass through the optimizer (channel) to produce control signals that influence the system (destination). In traditional MPC, a human designer establishes fixed statistical dependencies between states, controls, and responses—analogous to a communication system designed for expected noise conditions. When operating conditions change, these fixed relationships no longer effectively transmit the "message" of desired behavior, requiring manual redesign.

In contrast, an EL-enabled MPC system treats these statistical dependencies as dynamic. The Information Digital Twin (IDT) continuously monitors and calculates the mutual information $MI(S,A;S')$ between optimizer inputs $S$, actions $A$, and resulting states $S'$. By evaluating alignment through entanglement metrics, the IDT generates real-time adaptive signals to the optimizer. Using this information, the optimizer adapts control parameters to maintain structured dependencies as system dynamics shift. Unlike conventional MPC, EL ensures continuous model refinement without manual intervention.

### C. Entanglement as Bound Information

Entanglement Learning (EL) introduces the concept of bound information to describe dynamically interlocked relationships within control systems. Bound information refers to information that is constrained by its dependencies on other variables, meaning a change in one variable requires corresponding adjustments in the others to maintain the system's coherence. This concept extends beyond mutual information, which quantifies static statistical dependencies between variables at a specific point in time but does not capture the need for ongoing mutual adaptation when conditions change.

In Model Predictive Control (MPC), the states $S$, control actions $A$, and next states $S'$ are intricately linked by design, forming a system where a change in one component—like a new control action—requires the others to adjust to maintain performance. Consider a UAV navigating shifting winds. Mutual information (MI) between actions $A$ and responses $S'$ measures their correlation at a given moment; if the wind changes, MI simply updates to reflect the new relationship, offering no guidance on adaptation. By contrast, entanglement—here, a measure of how tightly $S$, $A$, and $S'$ co-evolve per the MPC model—detects when this relationship deviates from expectations. For example, if a wind gust disrupts the predicted link between thrust and altitude, entanglement signals the controller to adapt, ensuring system coherence.

In the context of EL, entanglement thus serves as more than just a measurement—it acts as a mechanism for ensuring system resilience. By quantifying bound information, the Information Digital Twin (IDT) continuously monitors the alignment between system components. When deviations occur, it generates adaptive signals, prompting the optimizer to adjust control strategies before performance deteriorates.

### D. Entanglement Metrics as Performance Indicators

Having established bound information as the foundation of Entanglement Learning, we now turn to its practical implications. Entanglement metrics, derived from this concept, serve as real-time indicators of system performance. These metrics quantify the strength and quality of bound information relationships between system components, providing a universal measure of control effectiveness that transcends domain-specific performance indicators.

These metrics—base entanglement (measuring the overall strength of bound information), asymmetry (identifying which component of the bound relationship is weakening), and memory (quantifying the persistence of information binding over time)—collectively assess the system's inter-dependencies across states, actions, and outcomes. Together, they create an information-theoretic signature that both detects when bound relationships weaken and diagnoses which specific relationships require reinforcement. This enables targeted adaptation strategies that maintain system coherence despite changing conditions, often before traditional performance indicators reveal problems. The following sections formalize these metrics mathematically and demonstrate their theoretical role in guiding specific MPC parameter adjustments.

## IV. ENTANGLEMENT LEARNING FRAMEWORK

The EL framework applies information-theoretic principles to quantify the mutual predictability of input states, actions, and resulting states of the system [29] [30]. Relying on information-based metrics, the EL evaluates the effectiveness of the optimizer in the context of MPC to transform its inputs—such as system model predictions, constraints, and objectives—into control actions that achieve desired outcomes. These metrics, referred to as entanglement metrics, provide the MPC with a universal self-reference mechanism for detecting performance deviations, enabling real-time adjustments with minimal human intervention.

The EL framework (Fig. 3) integrates with MPC via an IDT that operates alongside the optimizer. The IDT continuously monitors three critical variables: the optimizer's input states (predicted states, costs, and constraints), the control actions it generates, and the resulting states after these actions.

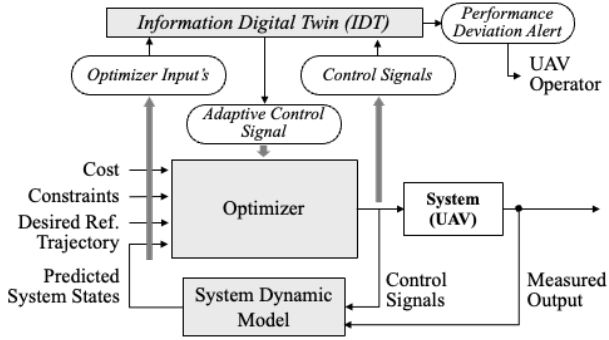

Fig. 3. Schematic representation of the MPC architecture, depicting the flow of optimizer inputs (predicted states, cost evaluations, current constraints, and desired trajectory) and outputs as control commands.

By computing entanglement metrics from the probability distributions of these variables, the IDT identifies misalignments between predicted and actual behaviors, generating adaptive signals to adjust MPC parameters while preserving the core functions of the optimizer.

### A. Entanglement Metrics

The entanglement metrics quantify different aspects of information relationships within the MPC system, measured in bits. These metrics derive from entropy, which measures uncertainty in a random variable as:

$$H(X) = -\sum p(x) \log^2 p(x) \qquad (1)$$

Where $p(x)$ is the probability of each possible value. Higher entropy indicates greater uncertainty, while mutual information between variables represents uncertainty reduction. Figure 4 illustrates these entropy relationships as a Venn diagram of three entropy regions: $H(S)$ for system states, $H(A)$ for control actions, and $H(S')$ for resulting states.

Figure 4 introduces a Venn diagram that maps out the entropy relationships among initial system states $H(S)$, control actions $H(A)$, and resulting states $H(S')$. The overlapping regions in this diagram highlight the mutual information shared between these variables, revealing their statistical dependencies in a visually intuitive way.

Specifically, the central overlap, denoted as $MI(S,A;S')$, defines the base entanglement $\psi$, which quantifies how states and actions jointly influence future states. Additional overlaps, such as $MI(A;S')$ and $MI(S;A)$, capture asymmetry in entanglement, reflecting the directional contributions of

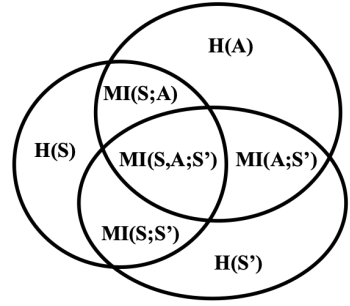

Fig. 4. Venn diagram showing entropy regions for states $H(S)$, actions $H(A)$, and resulting states $H(S')$, with overlaps indicating mutual information: $MI(S,A;S')$ (entanglement), $MI(A;S')$, $MI(S;A)$ (asymmetry), and $MI(S;S')$ (memory).

actions and states, while $MI(S;S')$ represents entanglement memory, linking past and future states. This structure goes beyond simple pairwise correlations, offering a multi-variable perspective on system coherence. Based on these entropy relationships, we define three complementary metrics that capture different aspects of controller-system alignment:

Base entanglement ($\psi$): Measures the mutual information between the combined state-action pair and the resulting state, quantifying the overall predictability of the control system:

$$\psi = I(S,A;S') = H(S,A) + H(S') - H(S,A,S') \qquad (2)$$

This metric evaluates the effectiveness of the optimizer's decisions in achieving the desired flight behavior. In MPC systems, entanglement represents how effectively the optimizer's knowledge of current states and chosen control actions predicts future system behavior. Higher entanglement indicates stronger alignment between the controller's model and actual UAV dynamics, while a decrease signals diminished control effectiveness, necessitating attention to prevent trajectory tracking degradation.

Entanglement asymmetry ($\Lambda\psi$): Quantifies the imbalance between different information pathways in the system:

$$\Lambda\psi = MI(A;S') - MI(S;A)$$

$$= [H(A) + H(S') - H(A,S')] - [H(S) + H(A) - H(S,A)] \qquad (3)$$

This metric diagnoses the root cause of performance issues, distinguishing whether they arise from discrepancies between the UAV model and actual flight dynamics, or from overly restrictive optimization constraints. Positive asymmetry indicates limitations in control execution, while negative asymmetry suggests errors in the prediction model. This insight directly informs corrective actions, such as updating dynamic model parameters or relaxing constraints to restore performance, enabling targeted interventions rather than general recalibration.

Entanglement memory ($\mu\psi$): Measures the natural predictability of the system independent of control actions:

$$\mu\psi = MI(S;S') = H(S) + H(S') - H(S,S') \qquad (4)$$

This metric identifies inconsistencies in the response of the UAV to control actions over time, often signaling environmental influences like wind patterns or shifts in vehicle dynamics. Memory captures how consistently system states evolve regardless of intervention, revealing underlying environmental dynamics. When memory decreases, it indicates that previously reliable patterns in system behavior are changing, which may require immediate updates to the system model to maintain performance.

Together, these metrics form a comprehensive information-theoretic signature of controller-system alignment. Deviations in these metrics, measured in bits, guide MPC adaptations based on their magnitude and pattern. Minor deviations prompt automatic adjustments to optimization weights and constraints, while significant changes may alert operators to the need for model reidentification or problem restructuring. The metrics work in concert to provide both detection and diagnosis—for example, increased asymmetry during wind disturbances might trigger updates to prediction model parameters, while declining memory could suggest extending the prediction horizon to address slower-evolving environmental effects. By quantifying information relationships rather than just tracking errors, the entanglement framework enables earlier and more targeted adaptation to evolving conditions.

### B. Entanglement Framework Hypothesis

The hypothesis underlying entanglement learning (EL) in MPC systems posits that the effectiveness of MPC can be systematically quantified by analyzing information patterns and gradients. These patterns measure the mutual predictability among the input state of the optimizer, its output control state, and subsequent input state, including UAV responses. Mathematically, we hypothesize that MPC performance correlates with entanglement metrics according to:

$$P(t) \propto f(\psi(t), \Lambda\psi(t), \mu\psi(t)) \qquad (5)$$

where P(t) represents controller performance (e.g., tracking accuracy, stability margins), and f is a monotonically increasing function of the entanglement metrics. Furthermore, we propose that temporal gradients in these metrics predict performance changes:

$$\Delta P(t+\tau) \propto g(\Delta\psi(t), \Delta\Lambda\psi(t), \Delta\mu\psi(t)) \qquad (6)$$

where τ represents a time delay between metric changes and observable performance effects, and g captures the sensitivity of performance to metric variations. Higher mutual predictability, referred to as entanglement, is hypothesized to correspond with enhanced trajectory tracking and system stability. If validated, these entanglement patterns could facilitate the IDT to detect performance degradation before it impacts flight safety. These predictive capabilities would enable timely model updates or optimization parameter adjustments while preserving the fundamental stability properties of the MPC.

## V. Defining the MPC Entanglement Model

Applying the entanglement learning framework to MPC requires a detailed understanding of the key information flows within the optimization process. The MPC optimizer converts inputs—derived from system dynamic model predictions, cost functions, and constraints—into control signals that direct system behavior. Entanglement metrics are defined by analyzing the probability distributions of these inputs, the resulting control actions, and subsequent model predictions. These metrics quantitatively evaluated the alignment between the optimizer's objectives and actual system behavior. This section defines these metrics for MPC systems, explains how to interpret their patterns, and demonstrates how they can inform specific parameter adjustments to maintain performance under changing conditions.

### A. MPC-EL Framework Integration

Monitoring the relationships between inputs and outputs of the MPC optimizer during operation is critical to assess its effectiveness. The optimizer processed multiple input streams, including predicted states from the system dynamic model, cost evaluations, current constraints, and the desired trajectory (see Fig. 3). These inputs collectively formed the input state of the optimizer. The outputs of the optimizer, recorded as control commands, constituted the action states of its system. These input and output states established the foundation for computing entanglement metrics.

The IDT evaluated MPC performance by analyzing the frequency of various optimizer input states and control actions across optimization cycles. It examined their mutual information patterns and gradients to construct EL metrics. These metrics, expressed in bits, characterized the behavioral patterns of the optimizer. The IDT quantified how effectively the optimizer aligned its decisions with UAV flight objectives by tracking these metrics and their trends over time.

### B. Interpreting MPC Performance Using EL Metrics

Entanglement ($\psi$) for MPC systems represents the predictability of desired outcomes based on a given input and its corresponding action. This is defined as:

$$Ent, \psi = MI \begin{pmatrix} Optimizer\ inputs, control\ actions; \\ Next\ Optimizer\ inputs \end{pmatrix} \qquad (7)$$

where MI denotes mutual information. Entanglement measures how effectively control actions achieve their intended outcomes. The system generates control actions from inputs, such as predicted states, costs, constraints, and desired trajectories, during each optimization cycle. These actions yield new inputs for the subsequent cycle, indicating whether the initial actions performed as expected. High entanglement signifies effective control actions, while reduced entanglement indicates performance issues.

Base entanglement reflects the MPC's ability to maintain control over UAV behavior. A decline suggests reduced control reliability, which may result from mismatches between the model and actual flight dynamics or overly restrictive optimization constraints. For instance, unpredictable wind patterns or reduced battery power can diminish control effectiveness, causing entanglement to decrease before trajectory tracking noticeably deteriorates. This early warning enables proactive adaptation.

When base entanglement decreases below established thresholds, the IDT could potentially generate adaptive signals to adjust key optimizer parameters. Such adjustments might target the prediction horizon and cost function weights. For example, the prediction horizon could be shortened to focus on more reliable near-term predictions:

$$Np_{adapted} = max(Np_{min}, Np_{nominal} \cdot (1 - \beta \cdot (\psi_{baseline} - \psi_{current})/\psi_{baseline})) \quad (8)$$

Where $\beta$ represents a sensitivity factor. Similarly, cost function weights could be rebalanced to emphasize more reliable control aspects, temporarily prioritizing objectives that maintain higher mutual information with system responses. These conceptual adjustment mechanisms illustrate how entanglement metrics might translate to specific parameter adaptations while maintaining system stability.

Entanglement asymmetry ($\Lambda\psi$) provides insights into the root cause of MPC performance problems and guides corrective actions. A significant increase in asymmetry often indicates that optimization constraints have become excessively restrictive, forcing the system to enforce rigid behaviors at the expense of adaptability, leading to inefficient or overly aggressive flight maneuvers. Conversely, minor asymmetry changes combined with declining entanglement suggest fundamental issues with the dynamic model. Such discrepancies may cause the optimizer to make reasonable decisions that fail to achieve the desired outcomes. These insights facilitate targeted interventions: adjusting optimization parameters to enhance flexibility or refining the dynamic model to better represent the current operating conditions of the UAV.

When asymmetry exceeds certain thresholds, the IDT could implement specific parameter adjustments based on its direction. Positive asymmetry might trigger constraint relaxation proportional to the measured deviation:

$$u_{bound,adapted} = (1 + \alpha \cdot max(0, \Lambda\psi)) \cdot u_{bound,nominal} \quad (9)$$

Where $\alpha$ determines adaptation sensitivity. This would allow the controller to temporarily exceed nominal input limits when necessary to maintain performance. Negative asymmetry could prompt updates to the dynamic model parameters, particularly those related to external forces:

$$c_{drag,adapted} = c_{drag,nominal} \cdot (1 - \gamma \cdot min(0, \Lambda\psi)) \quad (10)$$

where $\gamma$ scales the adaptation effect. These conceptual mechanisms illustrate how asymmetry patterns could inform precise adjustments targeting the specific source of misalignment, rather than general recalibration of the entire control system.

*Entanglement memory* ($\mu\psi$) evaluates the capacity of the system model to consistently represent the dynamic behavior of the UAV, independent of the control actions. High memory values indicate reliable state evolution predictions between the model and the actual flight dynamics. For example, a decline in memory underscores the necessity of updating the dynamic model to align with the current UAV behavior under new flight conditions. This metric is critical for identifying situations where model refinement is required, unlike adjustments in optimizer parameters to sustain overall performance.

When entanglement memory values decline below established thresholds, the IDT could trigger adaptations focused on the temporal aspects of the control problem. One potential adjustment involves modifying the sampling time of the controller to better capture evolving dynamics:

$$T_{s,adapted} = max(T_{s,min}, T_{s,nominal} \cdot (1 - \delta \cdot (\mu_{\psi,baseline} - \mu_{\psi,current})/\mu_{\psi,baseline})) \quad (11)$$

here $\delta$ represents a sensitivity parameter. Additionally, declining memory might suggest adjusting the state estimation filter parameters to better account for changing environmental conditions. This could involve increasing the process noise covariance matrix elements to give more weight to recent measurements over model predictions:

$$Q_{adapted} = Q_{nominal} \cdot (1 + \varepsilon \cdot (\mu_{\psi,baseline} - \mu_{\psi,current})/\mu_{\psi,baseline}) \quad (12)$$

These conceptual adaptations illustrate how memory metrics could inform specific adjustments to temporal parameters and state estimation components, helping the controller maintain alignment with changing system dynamics.

Although these metrics assess distinct aspects of MPC performance—overall predictability, degradation sources, and model fidelity—they exhibit important interdependence. This interrelationship enables the IDT to isolate performance issues and recommend targeted interventions. Figure 5 [31], illustrates entanglement metrics during RL agent training, demonstrating how information theory characterizes control system dynamics. The varying patterns throughout the learning process—from initial exploration through policy development—show how mutual information and entropy naturally capture the evolving state-action relationships. For MPC-based UAV controllers, analogous information patterns would characterize the controller-aircraft dynamics, providing a mathematical framework for monitoring system behavior; deviations from established baseline profiles would indicate misalignment requiring adaptation.

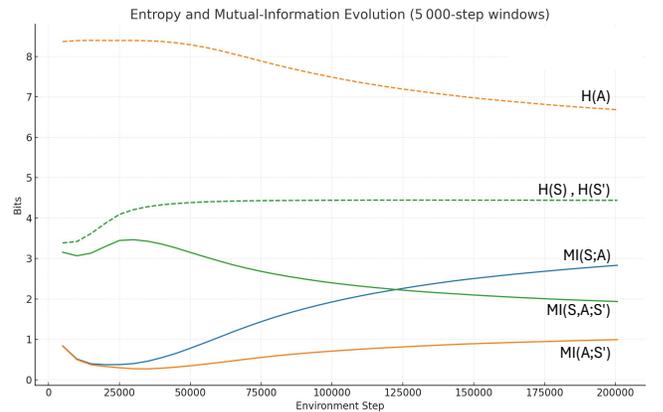

Fig. 5. Information-theoretic characterization of RL agent learning dynamics over 200k training steps, illustrating how entropy and mutual

information metrics capture evolving control behavior—a principle applicable to MPC-based control systems.

## C. Discretization and Computational Feasibility

The framework discretized the continuous variables of the MPC into finite bins to facilitate computing entanglement metrics, enabling the calculation of probability distributions. Optimizer inputs, including predicted states, cost values, constraints, and desired trajectories, are binned using adaptive ranges aligned with operational significance. Predicted states are grouped according to UAV flight envelope limits, cost values are binned near typical thresholds, and constraints are categorized around activation boundaries. Control actions are discretized based on actuator resolution and physical limitations. This binning strategy prioritized resolution in critical ranges where precision is essential while employing broader bins in extreme ranges to enhance computational efficiency.

The computational feasibility of the framework can be achieved through strategic optimization of both the algorithm's efficiency and the system's resource allocation. The Information Digital Twin (IDT) would operate on a dedicated processor, handling probability updates and calculating entropy and mutual information without burdening the Model Predictive Control (MPC) processor. This separation ensures that the MPC can focus on its primary optimization tasks without performance degradation.

Computationally, the entanglement calculations scale linearly with the number of discretized bins, O(N), where N is typically 40-50 per dimension. For a three-dimensional system, this results in 120-150 bins total, requiring only basic arithmetic and logarithm operations. On modern embedded processors, this translates to approximately 5-10 microseconds per calculation cycle—far less than the millisecond-scale MPC optimization cycle.

The framework minimizes computational overhead through incremental probability updates, adjusting only the bins affected by new data rather than recalculating entire distributions. This approach reduces computation time by an order of magnitude compared to naive implementations. Memory requirements remain modest, typically under 10 kilobytes for a system with 40-50 bins per dimension, encompassing storage for probability counters (40-50 floating-point values per distribution) and a sliding window of recent state-action pairs. The binning operation adds negligible overhead to the state sampling already performed in standard MPC implementations, ensuring minimal additional computational load.

This separation of responsibilities allows the IDT to function in parallel with the MPC system, maintaining real-time compatibility while preserving the integrity of core optimization operations. The computational burden of entanglement calculations accounts for less than 1% of the CPU resources required for a typical MPC optimization cycle, ensuring the approach remains practical for resource-constrained embedded systems in UAV applications.

## VI. IMPLEMENTING THE EL FRAMEWORK: THE INFORMATION DIGITAL TWIN

The IDT was a parallel monitoring system for an MPC controller, implementing the entanglement framework to evaluate and maintain system performance. Similar to conventional digital twins replicating physical systems, the IDT created a real-time information model of MPC performance by analyzing relationships among states and actions. The IDT comprised six primary modules, which collectively measure, analyze, and preserve the functionality of the MPC using information-theoretic principles.

1. Input processing: This module continuously monitored the MPC system, capturing three critical data streams: optimizer inputs (predicted states, costs, constraints, and desired trajectory), implemented control actions, and subsequent system responses. The module normalized and organized these data during each optimization cycle for probability distribution calculations, maintaining sliding windows of recent data to support real-time analysis.

2. EL metric calculations: This module computed the three core entanglement metrics using processed data streams. Probability distributions of input states and control actions over recent optimization cycles are used to calculate entanglement ($\psi$) for overall predictability, asymmetry ($\Lambda\psi$) for performance imbalances, and memory ($\mu\psi$) for model accuracy. These calculations operated simultaneously with the MPC system, ensuring no disruption to control timing.

3. EL baseline definition: Reference values for entanglement metrics are established during periods of verified optimal performance. The model characterized typical metric variations across operating conditions and developed adaptive thresholds to accommodate expected changes. The baseline evolved with accumulated operational data, providing a refined reference for identifying significant deviations.

4. Misalignment detection: Current entanglement metrics are continuously compared with baseline values to identify substantial deviations. The module evaluated the magnitude, patterns, and gradients of EL metrics, applying adaptive thresholds to consider various operational phases. Deviations exceeding these thresholds initiated deeper performance analyses while filtering out normal variations.

5. Alignment restoration analysis: The module diagnosed issues by examining relative changes in entanglement, asymmetry, and memory. It identified whether deviations resulted from model inaccuracies, optimization deficiencies, or environmental factors. Both immediate and long-term trends are analyzed to recommend targeted adaptation strategies.

6. Feedback generation: Two types of outputs are generated based on the analysis: adaptive signals for minor deviations, which modified MPC parameters (e.g., prediction horizons or constraint weights), and detailed alerts for significant issues requiring system-level updates. Feedback was prioritized based on the impact and urgency of the performance deviations, enabling timely and effective interventions.

The IDT framework can be validated in a high-fidelity simulation environment replicating UAV dynamics and MPC operation. Initial testing involved controlled variations, such as changes in wind conditions or vehicle parameters, to assess the sensitivity of the entanglement metric and the effectiveness of the adaptative responses. The simulation comprehensively evaluated automated adjustments and operator alerts without risking the physical systems. Furthermore, it provided a controlled setting to systematically validate stability guarantees during adaptation, defining performance limits before real-world deployment.

## VII. EL Potential Applications and Impact

Although EL remains a theoretical construct awaiting empirical validation, its foundation in Shannon's information theory provides a robust mathematical framework for measuring and optimizing system behavior. EL is applicable beyond UAV control, extending to any MPC implementation requiring adaptive performance. For example, autonomous vehicles can use EL to monitor trajectory optimization under diverse driving conditions, while industrial process control can maintain optimal performance as systems age or operating conditions evolve. EL offers a standardized method for quantifying and adapting controller behavior by introducing universal metrics expressed in bits. Rigorous testing across application domains is crucial to validate its utility and develop practical implementation guidelines, including distinguishing performance shifts caused by environmental factors from those arising from sensor degradation.

## VIII. Conclusion

This study introduced a novel framework for enhancing adaptability in Model Predictive Control (MPC) systems. Unlike existing methods focused on individual components, our approach uses information theory to create universal metrics, monitoring and maintaining MPC performance under changing conditions. Implemented through a non-invasive Information Digital Twin (IDT), it enables real-time adaptation in existing systems without affecting stability or architecture. This marks a key advance in autonomous control, optimizing performance across diverse, dynamic environments.

## IX. Future work

This study established a conceptual foundation for integrating EL as a parallel monitoring and adaptation layer within MPC systems. The IDT enhances the adaptability of existing MPC architectures in a non-invasive architecture-preserving manner, avoiding the need for fundamental redesigns of control algorithms. Future research will prioritize validating the framework through empirical studies in UAV control applications, extending its applicability to other MPC variants, such as distributed and stochastic MPC, and optimizing computational efficiency for real-time deployment in resource-constrained environments.

The proposed transition roadmap includes the following: (1) Validation in high-fidelity UAV simulations. (2) Hardware-in-the-loop testing with flight controllers. (3) Controllerflight tests within restricted airspace. (4) Gradual deployment in complex autonomous operations. Each phase will evaluate the impact of EL on MPC adaptability, stability, and control performance while developing practical implementation guidelines.